\documentclass[nonacm]{acmart}
\AtBeginDocument{%
  }

\newcommand\blfootnote[1]{%
  \begingroup
  \renewcommand\thefootnote{}\footnote{#1}%
  \addtocounter{footnote}{-1}%
  \endgroup
}

\usepackage{lipsum}
\usepackage{multirow}
\usepackage{soul}

\usepackage{amsmath}

\DeclareMathOperator{\bind}{\otimes}

\DeclareMathOperator{\bundle}{\oplus}

\newcommand{\continuousEncoder}{\mathcal{E}}

\newcommand{\vectorInverter}{\mathcal{I}}

\newcommand{\vectorNormalizer}{\mathcal{N}}

\newcommand{\cleanup}{\mathcal{C}}

\newcommand{\cleanupResonator}{\cleanup_{\hspace{1pt}\text{R}}}

\newcommand{\cleanupHopfield}{\cleanup_{\hspace{1pt}\text{H}}}

\newcommand{\codebook}{\Phi}

\newcommand{\vectorSim}{\mathcal{S}}

\newcommand{\spaceEnc}{\bigotimes_{j=1}^{n} \continuousEncoder_j(x_j)}

\newcommand{\spaceEncComp}{n\,O(\continuousEncoder)+(n-1)\,O(\bind)}

\newcommand{\valueEnc}{\continuousEncoder(v)}

\newcommand{\valueEncComp}{O(\continuousEncoder)}

\newcommand{\memEnc}{\mathbf{m} \bundle \big( \phi_p(\mathbf{x})\bind\phi_v(v) \big)}

\newcommand{\memEncComp}{O(\bind) + O(\bundle)}

\newcommand{\spatialInv}{\vectorNormalizer{(\mathbf{m})}\bind\vectorInverter\big(\phi_p(\mathbf{x}')\big)}

\newcommand{\spatialInvComp}{O(\phi_p(\mathbf{x}')) + O(\vectorInverter) +O(\bind)}

\newcommand{\resonator}{\vectorNormalizer\!\left(\bigoplus_{i=1}^{k}\codebook_i\bowtie\,\vectorSim\!\big(\phi_v(v'), \codebook_i\big)\right)}

\newcommand{\resonatorComp}{O(\vectorNormalizer)+k\big(O(\bowtie)+O(\vectorSim)\big)+(k-1)O(\bundle)}

\newcommand{\hopfield}{\vectorNormalizer\big(\bigoplus_{i=1}^{k}\Phi_i\bowtie\,\vectorNormalizer_\text{H}\big(\mathcal{S}(\phi_v(v'), \Phi_i)\big)\big)}

\newcommand{\hopfieldComp}{k\big(O(\bowtie)+O(\vectorSim)\big) +O(\vectorNormalizer_\text{H}) + O(\vectorNormalizer)+(k-1)O(\bundle)}

\newcommand{\regressSingle}{\sum_{i=1}^k \varphi_i \sigma_i\Big(
        \beta\vectorSim\big(\Phi_i,\phi_v(v')\big)
    \Big)
}

\newcommand{\regressSingleComp}{k\,O(\vectorSim)+\epsilon}

\newcommand{\regressNeural}{f_\theta(\phi_v(v'))}

\newcommand{\regressNeuralComp}{\epsilon}

\begin{document}

\title{HyperSpace: A Generalized Framework for Spatial Encoding in Hyperdimensional Representations}

\author{Shay Snyder}
\orcid{0000-0002-3369-3478}
\affiliation{%
  \institution{George Mason University}
  \city{Fairfax}
  \state{Virginia}
  \country{USA}
}
\email{ssnyde9@gmu.edu}

\author{Andrew Capodieci}
\orcid{0009-0007-1272-6465}
\affiliation{%
  \institution{Neya Systems}
  \city{Pittsburgh}
  \state{Pennsylvania}
  \country{USA}}
\email{acapodieci@neyarobotics.com}

\author{David Gorsich}
\orcid{0000-0003-2961-1393}
\affiliation{%
  \institution{U.S. Army Ground Systems}
  \city{Warren}
  \state{Michigan}
  \country{USA}
}
\email{david.j.gorsich.civ@army.mil}

\author{Maryam Parsa}
\orcid{0000-0002-4855-4593}
\affiliation{%
 \institution{George Mason University}
 \city{Fairfax}
 \state{Virginia}
 \country{USA}}
 \email{mparsa@gmu.edu}

\renewcommand{\shortauthors}{Snyder et al.}

\begin{abstract}
Vector Symbolic Architectures (VSAs) provide a well-defined algebraic framework for compositional representations in hyperdimensional spaces.
However, evaluating their behavior on continuous spatial domains remains challenging due to the diversity of backends and the absence of a unified software framework for systematic evaluation.
We introduce \textit{HyperSpace}, an open-source framework that decomposes VSA systems into modular operators for encoding, binding, bundling, similarity, cleanup, and regression.
Using HyperSpace, we analyze and benchmark two representative VSA backends: Holographic Reduced Representations (HRR) and Fourier Holographic Reduced Representations (FHRR).
Although FHRR provides lower theoretical complexity for individual operations, HyperSpace’s modularity reveals that similarity and cleanup dominate runtime in spatial domains.
As a result, HRR and FHRR exhibit comparable end-to-end performance.
Differences in memory footprint introduce additional deployment trade-offs where HRR requires approximately half the memory of FHRR vectors.
By enabling modular, system-level evaluation, HyperSpace reveals practical trade-offs in VSA pipelines that are not apparent from theoretical or operator-level comparisons alone.
\end{abstract}




\received{8 April 2026}

\maketitle

\section{Introduction}
\label{section:introduction}

Vector Symbolic Architectures (VSAs), also commonly known as hyperdimensional computing, provide an algebraic framework for representing and manipulating structured information with high-dimensional distributed vectors~\cite{kanerva1994spatter, Kleyko_2022}.
Rather than relying on learned latent spaces, VSAs construct compositional representations through operations such as binding, bundling, inversion, and similarity~\cite{Kleyko_2022, plate1991holographic}.
These operations support associative recall and symbolic composition within a unified mathematical framework~\cite{imani2017exploring, plate2003holographic}.\blfootnote{DISTRIBUTION STATEMENT A. Approved for public release; distribution is unlimited. OPSEC \#10534.}

Recent work has shown that VSAs support a broad range of capabilities, including generative modeling, probabilistic inference, graph learning, and associative memory~\cite{furlong2024modelling, nunes2022graphhd, frady2020resonator}.
VSAs have also shown growing promise for spatial reasoning problems, where continuous coordinates, maps, and value functions must be represented with compositional memory operations~\cite{KomerBrent2020, dumont2025symbols, frady2022computing}.
This capability is relevant for robotics and autonomous systems, where spatial representations support mapping, navigation, path planning, and continuous decision-making over structured environments~\cite{snyder2026brain, snyder2025generalizable, reda2024path, Renner2024}.

However, evaluating VSA systems on continuous spatial domains remains difficult.
Existing spatial VSA implementations are often developed in task-specific settings, making it hard to separate the effects of the underlying backend from the effects of encoding strategy, cleanup dynamics, decoder design, or evaluation protocol~\cite{torchhd, bekolay2014nengo, Renner2024, snyder2026brain}.
This limitation becomes especially important in dense pipelines, where end-to-end performance depends not only on the theoretical cost of individual operators, but also on how those operators interact across encoding, storage, inversion, cleanup, and regression.
As a result, isolated operator-level comparisons can fail to predict runtime, memory usage, and reconstruction behavior in full spatial workflows.

To address this gap, we introduce \textit{HyperSpace}\footnote{Publically available at \url{https://github.com/Parsa-Research-Laboratory/HyperSpace}}, an open-source framework for constructing and evaluating VSA pipelines with continuous spatial representations.
Rather than committing to a single backend, HyperSpace decomposes spatial processing into a shared set of abstract modules for positional encoding, value encoding, memory storage, positional inversion, cleanup, and regression.
This abstraction enables heterogeneous VSA backends to be instantiated within the same end-to-end pipeline and compared under controlled conditions.
As a result, HyperSpace serves both as an implementation framework and as a methodological tool for identifying computational, memory, and representational trade-offs that are not apparent when evaluating isolated operators.

Using HyperSpace, we analyze two representative VSA backends: Holographic Reduced Representations (HRR) and Fourier Holographic Reduced Representations (FHRR)~\cite{plate1991holographic, plate2003holographic}.
Although FHRR offers lower theoretical complexity for individual operations, our system-level analysis shows that these savings do not necessarily translate into lower end-to-end cost in spatial domains.
Instead, similarity-based cleanup and regression dominate runtime across many configurations, yielding comparable overall latency between HRR and FHRR.
At the same time, differences in vector representations create deployment-dependent memory trade-offs, with floating point HRR requiring approximately half the storage of complex FHRR vectors.
These results demonstrate why backend selection in spatial VSA systems should be treated as a pipeline-level design decision rather than a comparison of isolated operators.

The contributions of this work are summarized as follows:
\begin{itemize}
    \item We introduce HyperSpace, an open-source framework for constructing and evaluating continuous spatial VSA pipelines through shared abstract operators.
    \item We formalize spatial VSA processing as a sequence of modules, enabling controlled comparisons across heterogeneous backends, cleanup mechanisms, and regression methods.
    \item We instantiate HyperSpace with representative HRR- and FHRR-based backends and analyze their computational and memory complexity across the full pipeline.
    \item We identify deployment-dependent trade-offs between HRR and FHRR, showing that backend selection depends jointly on runtime, memory footprint, and decoder configuration.
\end{itemize}
\section{Background \& Previous Works}
\label{section:background}

VSAs represent information using high-dimensional vectors and distributed algebraic operations~\cite{kanerva1994spatter}.
Fundamental VSA operations include binding, bundling, similarity, and inversion~\cite{Kleyko_2022}.
Binding combines vectors into compositional representations, bundling superposes multiple items into a shared memory, similarity supports associative retrieval, and inversion enables partial recovery of bound vectors.
Together, these operations provide an algebraic framework for representing information within distributed vectors.

Multiple VSA backends have been proposed, including Holographic Reduced Representations (HRR)~\cite{plate1991holographic}, Fourier Holographic Reduced Representations (FHRR)~\cite{plate2003holographic}.
These backends differ in their algebraic formulations, numerical representations, and implementation costs, but they expose closely related operations for encoding, storage, and retrieval~\cite{Kleyko_2022}.
As a result, many VSA systems can be described at the level of shared abstract operations, even when their concrete implementations differ substantially.

\textit{VSA-based Spatial Representations: }
VSAs have increasingly been used to represent spatial structure in robotics, navigation, and cognitive modeling~\cite{dumont2025symbols, snyder2026brain, Renner2024, KomerBrent2020}.
In particular, spatial representations such as Spatial Semantic Pointers~\cite{KomerBrent2020, dumont2025symbols} and fractional power codes~\cite{frady2022computing, Renner2024} encode continuous coordinates into high-dimensional vectors that preserve geometric relationships while remaining compatible with existing VSA operators.
This makes VSAs attractive for tasks such as spatial memory, path planning, and continuous value representations~\cite{frady2022computing, Renner2024, KomerBrent2020, dumont2025symbols, snyder2026brain}.
However, spatial VSA systems are typically evaluated within task-specific implementations, making it difficult to separate backend properties from choices in encoding strategy, cleanup dynamics, decoder design, or evaluation protocol.
This challenge is particularly important in continuous spatial domains, where end-to-end performance depends not only on the cost of individual VSA operators, but also on how those operators interact across encoding, storage, cleanup, and decoding.

\textit{VSA Evaluation Frameworks: }
Existing software libraries such as TorchHD~\cite{torchhd} and Nengo-SPA~\cite{bekolay2014nengo} provide the infrastructure for constructing and experimenting with VSA backends.
However, these libraries primarily emphasize representation building and model development rather than controlled, system-level comparisons of heterogeneous VSA backends operating on a common workflow.
Consequently, prior works compare individual operators on isolated downstream tasks, making it difficult to determine which trade-offs arise from the backend or the surrounding pipeline~\cite{frady2020resonator, KomerBrent2020, dumont2025symbols}.

To address this gap, we introduce \textit{HyperSpace}, a modular framework for analyzing spatial VSA systems through a shared set of abstract operators.
Rather than committing to a single backend, HyperSpace decomposes the pipeline into continuous encoding, binding, bundling, inversion, similarity, cleanup, and regression modules that can be instantiated by different backends.
This abstraction enables comparisons of heterogeneous backends within a unified environment.
As a result, HyperSpace serves not only as an implementation framework, but also as a tool for identifying computational, memory, and representational trade-offs that are not apparent when evaluating isolated VSA operators.
\section{The HyperSpace Framework}
\label{section:research_methods}

\begin{figure*}[ht]
    \centering
    \includegraphics[width=0.90\linewidth]{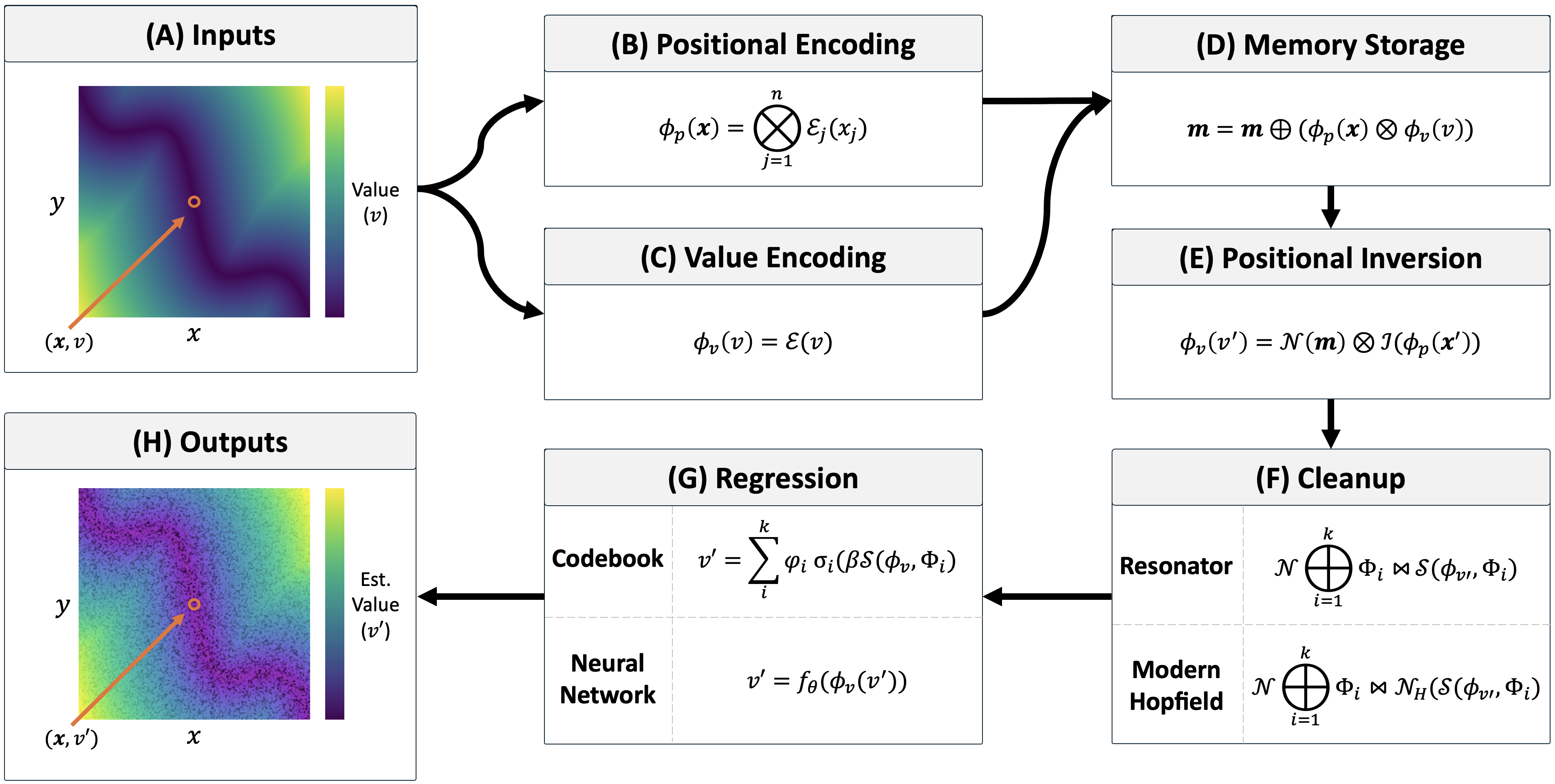}
    \caption{\textbf{
        A high-level overview of the HyperSpace framework.
        (A) Inputs consist of coordinate–value pairs $(\mathbf{x}, v)$.
        (B) Coordinates are encoded into hypervectors $\phi_p(\mathbf{x})$ via compositional positional encoding.
        (C) Values are encoded as hypervectors $\phi_v(v)$.
        (D) Position–value pairs are bound and bundled into a shared memory $m$.
        (E) Querying is performed by positional inversion and unbinding.
        (F) Retrieved representations are refined via cleanup operations (resonator or modern Hopfield).
        (G) A regression operator decodes the representation into a scalar estimate.
        (H) The final output yields reconstructed values $v'$ at query locations.
    }}
    \Description{A high-level overview of the HyperSpace framework.}
    \label{fig:framework-overview}
\end{figure*}

\noindent
HyperSpace creates an abstract framework for processing continuous spatial information in VSAs through a series of extensible modules.
The framework is agnostic to any specific VSA implementation and instead formalizes a minimal set of abstract operations:
\begin{itemize}
    \item \textbf{Continuous Encoding} ($\continuousEncoder$): maps continuous spatial coordinates or scalar values into high-dimensional vectors
    \item \textbf{Binding} ($\bind$): combines two or more high-dimensional vectors into a compositional representation
    \item \textbf{Bundling} ($\bundle$): aggregates multiple vectors into a composite representation
    \item \textbf{Similarity} ($\vectorSim$): computes the correlation between vectors for retrieval and decoding
    \item \textbf{Inversion} ($\vectorInverter$): computes the inverse of a vector, enabling the retrieval of constituent components from composite representations
    \item \textbf{Normalization} ($\vectorNormalizer$): normalizes the vector to unit length
    \item \textbf{Cleanup} ($\cleanup$): iteratively remove noise from output vectors
    \item \textbf{Weighting} ($\bowtie$): modulate the contribution of a vector by a scalar weight
    \item \textbf{Regression} ($\mathcal{R}$): abstract function for converting from vector space to real-value space
\end{itemize}
A high-level overview of the framework with the operational implementation of each module, as a function of these abstract operations, is shown in Figure~\ref{fig:framework-overview}.
These abstractions allow complexity to be expressed generically in terms of the number of encoding, binding, bundling, similarity, and inversion operations, independent of the underlying algebraic formulation.
Throughout the remainder of the manuscript, we refer to this type of complexity as \textit{Operational Complexity}. The operational implementation and complexity of each module is shown in Table~\ref{table:operational-implementation-and-complexity}.
Moreover, all mathematical symbols leveraged throughout the manuscript are detailed in Table~\ref{tab:symbol-table}.
The following subsections describe each HyperSpace module.

\textbf{Input Representations:} 
To provide a modular framework for processing continuous spatial information, HyperSpace assumes that each dataset $\mathcal{D}$ consists of $N$ samples:
\begin{equation}
    \label{equation:input-format}
    \mathcal{D} = \left\{ (\mathbf{x}_i, v_i) \;\middle|\; 
    \mathbf{x}_i \in \mathbb{R}^n,\; v_i \in \mathbb{R} \right\}_{i=1}^{N},
\end{equation}
where each sample is defined by a position vector $\mathbf{x}$ representing an $n$-dimensional spatial location, and an associated scalar value $v$. As illustrated in the example shown in Figure~\ref{fig:framework-overview}(A), each sample $(\mathbf{x}, v)$ 
corresponds to a position vector $\mathbf{x} = (x, y)$ within a 2D environment and an associated scalar value $v$, representing the measurement at that location.

{
\renewcommand{\arraystretch}{1.7}
\begin{table*}[]
\centering
\caption{The baseline implementation of HyperSpace defined through abstract algebraic terminology can will be extended by downstream VSA systems.}
\begin{tabular}{l|l|l|l}
\hline
\textbf{Module}                          & \textbf{Variant} & \textbf{Abstract Definition} & \textbf{Operational Complexity}  \\ \hline
\multirow{1}{*}{Spatial Encoding}        & -    & $\spaceEnc$         & $\spaceEncComp$      \\ \hline
\multirow{1}{*}{Value Encoding}          & -    & $\valueEnc$         & $\valueEncComp$      \\ \hline
\multirow{1}{*}{Memory Storage}          & -    & $\memEnc$           & $\memEncComp$        \\ \hline
\multirow{1}{*}{Positional Inversion}    & -    & $\spatialInv$       & $\spatialInvComp$    \\ \hline
\multirow{2}{*}{Cleanup}                 & Resonator        & $\resonator$        & $\resonatorComp$     \\
                                         & Modern Hopfield  & $\hopfield$         & $\hopfieldComp$      \\ \hline
\multirow{2}{*}{Regression}              & Codebook         & $\regressSingle$    & $\regressSingleComp$ \\
                                         & Neural Network   & $\regressNeural$    & $\regressNeuralComp$ \\ \hline
\end{tabular}
\label{table:operational-implementation-and-complexity}
\end{table*}
}

\textbf{Positional Encoding:} 
As shown in Figure 1(B), the positional encoding module transforms each position vector 
$\mathbf{x}$ into a hypervector $\phi_p(\mathbf{x})$ by encoding each individual dimension of $\mathbf{x}$ 
and binding the resulting encodings together:
\begin{equation}
    \label{equation:spatial-encoding}
    \phi_p(\mathbf{x})\;=\;\spaceEnc,
\end{equation}
where $\mathcal{E}_j$ denotes the abstract encoder for the $j$-th component of $\mathbf{x}$, 
and $\otimes$ represents an abstract binding operator. 
The operational complexity of positional encoding $\phi_p(\mathbf{x})$ scales with the dimensionality of the point vector $n$. Therefore the operation requires $n$ encodings and $(n-1)$ bindings, 
yielding a total operational complexity of:
\begin{equation}
    \label{equation:spatial-encoding-complexity}
    O(\phi_p(\mathbf{x}))\;=\;\spaceEncComp.
\end{equation}
This formulation allows position vectors to be constructed through iterative or parallel binding and encoding operations.

\textbf{Value Encoding:} 
As shown in Figure 1(C), the value encoding module transforms each scalar input into a hypervector with Equation~\ref{equation:spatial-encoding} where $n=1$:
\begin{equation}
    \label{equation:value-encoding}
    \phi_v(v)\;=\;\valueEnc.
\end{equation}
Accordingly, the operational complexity of $\phi_v(v)$ is the simplified form of 
Equation~\ref{equation:spatial-encoding-complexity} with $n=1$:
\begin{equation}
    \label{equation:value-encoding-complexity}
    O(\phi_v(v))\;=\;\valueEncComp.
\end{equation}
Both $\phi_p(\mathbf{x})$ and $\phi_v(v)$ produce hypervectors in the same representation space, 
enabling compositional binding or superposition within downstream HyperSpace modules.

\textbf{Memory Storage:} 
As shown in Figure 1(D), the memory storage module aggregates positional and value hypervector pairs into a single memory vector $\mathbf{m}$. 
Each encoded position and value pair $(\phi_p(\mathbf{x}), \phi_v(v))$ is added to the memory by binding the constituent vectors together before bundling with the existing memory vector:
\begin{equation}
    \label{equation:memory-encoding}
    \mathbf{m}\;=\;\memEnc,
\end{equation}
where $\otimes$ denotes the binding operator and $\oplus$ denotes the bundling operator. 
For each sample in the dataset $\mathcal{D}$, the memory storage module requires one bind and one bundle, 
yielding a per-sample operational complexity of:
\begin{equation}
    \label{equation:memory-encoding-complexity}
    O(\mathbf{m})\;=\;\memEncComp.
\end{equation}
This formulation enables the construction of distributed associative memories where spatial and value information are jointly encoded within a unified vector representation.

\textbf{Positional Inversion:} 
As shown in Figure 1(E), the entire dataset $\mathcal{D}$ of spatially associated values is encoded into the hypervector memory $\mathbf{m}$, the value at any location $\mathbf{x}'$ can be estimated by normalizing $\mathbf{m}$ and performing an inverse binding operation to recover a noisy approximation of the corresponding value vector $\phi_v(v')$:
\begin{equation}
    \label{equation:spatial-inversion}
    \phi_v(v')\;=\;\spatialInv,
\end{equation}
where $\vectorNormalizer$ denotes the abstract vector normalization operator and $\vectorInverter$ denotes the abstract vector inversion operator. 
The spatial inversion module requires one positional encoding, one vector inversion, one normalization, and one binding operation per query location.
While normalization is required, it can be computed once and reused across multiple query locations, allowing its cost to be amortized and excluded from the per-query complexity.
Therefore, the total operational complexity per query is:
\begin{equation}
    \label{equation:spatial-inversion-complexity}
    O(\phi_v(v'))\;=\;\spatialInvComp.
\end{equation}

A critical distinction between positional inversion and traditional end-to-end resonator-based systems 
is that positional inversion explicitly preserves spatial structure through the encoding and inversion process. 
In classical resonator formulations~\cite{frady2020resonator, Renner2024}, 
the retrieval operation seeks convergence to an attractor basin rather than decoding at fixed query locations.
Consequently, these models can recall stored item associations but cannot reliably recover values at arbitrary, previously unseen positions. 
In contrast, HyperSpace’s positional inversion module enables value estimation at any location.

{
    \renewcommand{\arraystretch}{1.4}
    \begin{table}[ht]
    \begin{center}
    \caption{The mathematical symbols used throughout this manuscript.} 
    \label{tab:symbol-table}  
    \begin{tabular}{c|c}
    \hline
    \textbf{Symbol}           & \textbf{Meaning}                  \\ \hline
    $\continuousEncoder$      & continuous encoder                \\
    $\bind$                   & bind                              \\
    $\bundle$                 & bundle                            \\
    $\vectorSim$              & vector similarity                 \\
    $\vectorInverter$         & vector inversion                  \\
    $\vectorNormalizer$       & vector normalizer                 \\
    $\cleanup$                & cleanup                           \\
    $\bowtie$                 & vector weighting                  \\
    $\mathcal{R}$             & regression decoder                \\
    $f_\theta$                & neural network                    \\
    \hline
    $\mathcal{D}$             & dataset                           \\
    $\mathbf{x}$              & position vector                   \\
    $v$                       & value scaler                      \\
    $v'$                      & decoded value scaler              \\
    $n$                       & environment dimensionality        \\
    $N$                       & dataset size                      \\
    $i$                       & abstract iterator                 \\
    $\phi_p(\mathbf{x})$      & position hypervector              \\
    $\phi_v(v)$               & value hypervector                 \\
    $\mathbf{m}$              & hypervector memory                \\
    $\Phi$                    & codebook                          \\
    $\varphi$                 & codebook values                   \\
    $(\cdot)'$                & estimate of a variable            \\
    $|\cdot|$                 & cardinality                       \\
    $q$                       & individual quantization level     \\
    $t$                       & discrete time step                \\
    $\beta$                   & temperature                       \\
    \hline


    \end{tabular}
    \end{center}
    \end{table}
}

\textbf{Cleanup:} 
As shown in Figure 1(F), HyperSpace's cleanup module implements associative memory cleanup through a generalized cleanup operator $\cleanup$. This operator introduces temporal dynamics where decoded value vectors are iteratively "cleaned" for some number of timesteps. Given a decoded value vector $\phi_v(v')$, the cleanup operator is defined over one timestep $t$ as:
\begin{equation}
    \phi_v(v')^{(t+1)} = \cleanup(\phi_v(v')^{(t)}).
\end{equation}
These cleanup operations commonly leverage a codebook $\Phi$ that contains a set of vectors representing $\phi_v$ at discrete points. Therefore, a proper derivation of the cleanup operator is defined as
\begin{equation}
    \phi_v(v')^{(t+1)} = \cleanup(\phi_v(v')^{(t)},\Phi),
\end{equation}
where the operational complexity of $\cleanup$ will scale with the cardinality of the codebook $|\Phi|$.

HyperSpace provides two baseline cleanup operators: Resonator Networks~\cite{frady2020resonator} and Modern Hopfield Networks~\cite{ramsauer2020hopfield}. We refer to these cleanup operations as $\cleanup_{\hspace{1pt}\text{R}}$ and $\cleanup_{\hspace{1pt}\text{H}}$, respectively.

The resonator cleanup operator $\cleanupResonator$ performs iterative refinement through recurrent superposition and feedback with the codebook:
\begin{equation}
    \label{equation:resonator-update}
    \cleanupResonator(\phi_v(v'),\codebook)\;=\;\resonator,
\end{equation}
where $\vectorSim(\cdot,\cdot)$ denotes a similarity operator, $\bundle$ represents bundling, $\vectorNormalizer(\cdot)$ is an optional normalization operator, $\bowtie$ is a weighted normalization operator that modulates the strength of the given codebook element $\Phi_i$ by its similarity with the query vector, and $k=|\Phi|$ is the cardinality of the codebook. 
Through recurrent updates, $\cleanupResonator$ converges toward a stable attractor corresponding to the most similar stored code.
The operational complexity of $\cleanupResonator$, for a single timestep, is defined as:
\begin{equation}
    O(\cleanupResonator)\;=\;\resonatorComp.
\end{equation}

In contrast, Hopfield cleanup $\cleanupHopfield$ is implemented as:
\begin{equation}
    \label{equation:hopfield-update}
    \cleanupHopfield(\phi_v(v'),\codebook)\;=\;\hopfield,
\end{equation}
where $\mathcal{N}_\text{H}$ is the softmax function. 
This process corresponds to minimizing an implicit energy function whose minima coincide with stored attractors.
The operational complexity of the Hopfield cleanup operator is defined as:
\begin{equation}
    O(\cleanupHopfield)\;=\;\hopfieldComp
\end{equation}
The major takeaway here is that each $\cleanupHopfield$ step requires an additional normalization operation compared to $\cleanupResonator$. 

\textbf{Regression:} In the final HyperSpace module, we transform the cleaned query vectors $\phi_v(v')$ into an estimated value $v'\in\mathbb{R}$ with an abstract regression operator $\mathcal{R}$. As shown in Figure 1(G) HyperSpace provides two implementations which we refer to as codebook $\mathcal{R}_{\hspace{1pt}\text{C}}$ and neural network $\mathcal{R}_{\hspace{1pt}\text{N}}$ decoding.
Codebook decoding is defined as an expectation over the discretized codebook values $\varphi$ and the similarity between the query vector and the corresponding code:
\begin{equation} \label{equation:base_regres_soft_discrete}
    v'\;=\;\regressSingle,
\end{equation}
where $i$ is an iterator representing the $i$-th discrete value in the codebook. The operational complexity of codebook decoding is defined as
\begin{equation}
    \label{equation:codebook-complexity}
    O(\mathcal{R}_{\hspace{1pt}\text{C}}(\phi_v(v')))\;=\;\regressSingleComp,
\end{equation}
where $\epsilon$ representing the additional logit-wise operations with the softmax and expectation. Given that $k <<\text{size}(\phi_v)$ and the softmax is only used once, we assume this overhead negligible and won't count towards operational complexity.

Neural network decoding is defined as:
\begin{equation} \label{equation:base_regres_neural}
    v'\;=\;\regressNeural,
\end{equation}
where $f_\theta$ is an arbitrary neural network that returns a single scalar value representing the decoded value. Given that no VSA operations are included with this regression approach, we assume the operational complexity is zero. However, the additional training and inference latency will be included in our results.

\begin{table*}[t]
\centering
\caption{Asymptotic Complexity Comparison of HRR~\cite{plate1991holographic} and FHRR~\cite{plate2003holographic} Operations}
\label{tab:operation_complexity}
\begin{tabular}{lccl}
\toprule
\textbf{Operation} & \textbf{HRR}~\cite{plate1991holographic} & \textbf{FHRR}~\cite{plate2003holographic} & \textbf{Explanation} \\
\midrule
Encoding $\mathcal{E}$ & $O(D \log D)$ & $O(D)$ & FFT-based FPE vs. direct complex exponentiation \\
Binding $\otimes$ & $O(D \log D)$ & $O(D)$ & Circular convolution (FFT) vs. element-wise multiplication \\
Bundling $\oplus$ & $O(D)$ & $O(D)$ & Both use element-wise addition \\
Similarity $\mathcal{S}$ & $O(D)$ & $O(D)$ & Both compute dot products \\
Inversion $\mathcal{I}$ & $O(D \log D)$ & $O(D)$ & FFT + conjugate + IFFT vs. conjugation only \\
Normalization $\vectorNormalizer$ & $O(D)$ & $O(D)$ & Both use norm + element-wise multiplication \\
Weighting $\bowtie$     & $O(D)$ & $O(D)$ & Both use element-wise multiplication \\
\midrule
\textbf{Storage per vector} & \textbf{4D bytes} & \textbf{8D bytes} & float32 vs. complex64 representation \\
\bottomrule
\end{tabular}
\end{table*}








\section{Results \& Discussion}
\label{section:results}


\noindent
We use HyperSpace to evaluate the computational and memory characteristics of two representative VSA backends: Holographic Reduced Representations (HRR)~\cite{plate1991holographic} and Fourier Holographic Reduced Representations (FHRR)~\cite{plate2003holographic}.
We first present an algorithmic analysis that examines how backend-specific operations interact with each module.
We then describe the procedure used to generate synthetic training and validation data.
Finally, we report empirical results that quantify the performance of each backend and highlight deployment considerations that motivate future research.

\subsection{Algorithmic Analysis}
We analyze the theoretical computational and memory complexity of HRR and FHRR within HyperSpace.
Rather than advocating for a single backend, this highlights how each backend interacts with different workloads, batching strategies, and deployment constraints.

Table~\ref{tab:operation_complexity} shows the asymptotic complexity of each VSA operation in HRR and FHRR.
HRR performs binding and encoding with an additional Fast Fourier Transform (FFT), yielding $O(D \log D)$ complexity for hypervectors of dimensionality $D$, whereas FHRR operates directly in the frequency domain with $O(D)$ element-wise operations.
However, HyperSpace performs all VSA operations in batched operations, where FFT costs amortize across multiple data samples.
As batch size increases, FFT complexity is minimized and HRRs approach linear scaling in $D$.


Beyond arithmetic complexity, memory complexity introduces additional considerations. HRR requires $4D$ bytes per vector (float32), whereas FHRR requires $8D$ bytes for complex values.
In distributed or communication-constrained environments, memory bandwidth, cache utilization, and inter-agent transmission costs may constrain overall system performance~\cite{10.1145/3724129}.
Therefore, HRR’s reduced memory footprint may outweigh FHRR’s arithmetic advantage.


This systems-level perspective is a key finding enabled by HyperSpace.
When evaluated within a complete pipeline, rather than isolated algorithmic primitives, HRR and FHRR exhibit comparable scaling behavior.
Consequently, backend selection shifts from a purely algorithmic comparison to a deployment-dependent trade-off.
FHRR provides streamlined frequency-domain computation with consistent $O(D)$ element-wise operations, whereas HRR offers storage and memory efficiency.
These findings suggest that backend choice is context-dependent, motivating future exploration of hybrid, quantized, or adaptive VSA representations that dynamically balance computational and memory efficiency.

\subsection{Experimental Analysis}

\vspace{5pt}
\noindent
\textbf{Data Generation:} We generate synthetic 2D maps to simulate cost maps commonly used in robotics~\cite{reda2024path}.
Each map is initialized as a continuous grid with start and goal positions at opposing corners.
The lowest cost path is defined by fitting a cubic spline through seven control points between the start and goal.
The spline is sampled at 10{,}000 points to approximate a continuous path.
All spline samples are assigned a uniform cost of $1.0$, thereby defining the navigable path.
The full map is constructed by computing the Euclidean distance transform (EDT) from the optimal path~\cite{2020SciPy-NMeth}.
The EDT induces a radially increasing cost structure where traversal cost grows proportionally with distance from the reference trajectory.
This produces smooth gradients that penalize deviation from the optimal path.
We sample the grid at a fixed resolution of $28\times28$ to mimic the scale of MNIST~\cite{lecun2010mnist} and N-MNIST~\cite{nmnist}.
This discretization results in $28 \times 28 = 784$ samples per map.

\begin{figure*}
    \centering
    \includegraphics[width=0.95\linewidth]{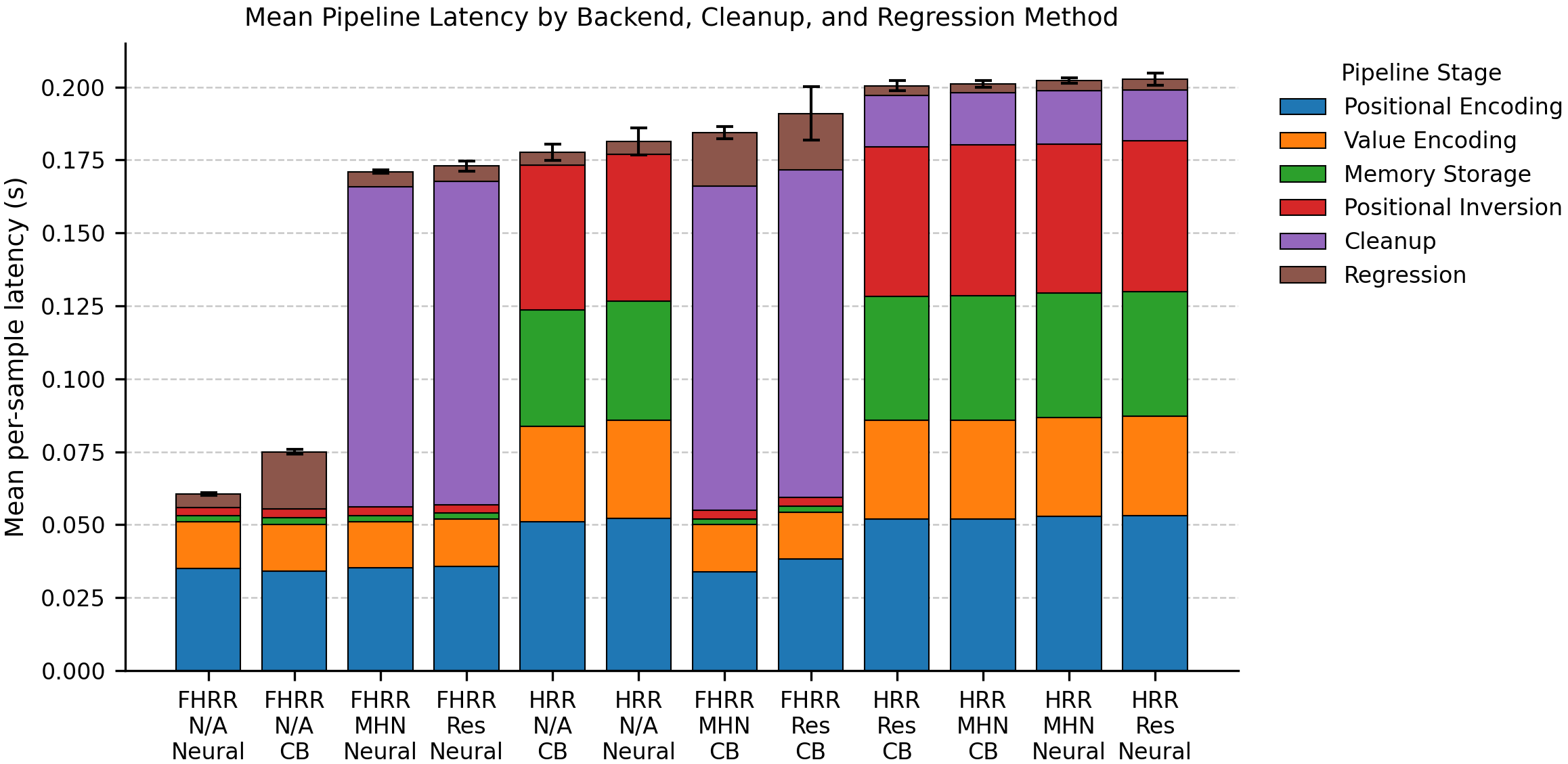}
    \caption{
        Pipeline latency breakdown for HRR and FHRR backends. Stacked bars show the mean per-stage latency, averaged over five random seeds. Abbreviations: MHN = Modern Hopfield Network; Res = Resonator Network; CB = codebook decoding.
    }
    \label{fig:pipeline-analysis}
    \Description{
        Pipeline latencies with HRR and FHRR. The stacked bars show the mean latency for each pipeline stage. Results are averaged across 5 seeds for each configuration.
    }
\end{figure*}

\begin{figure}
    \centering
    \includegraphics[width=0.7\linewidth]{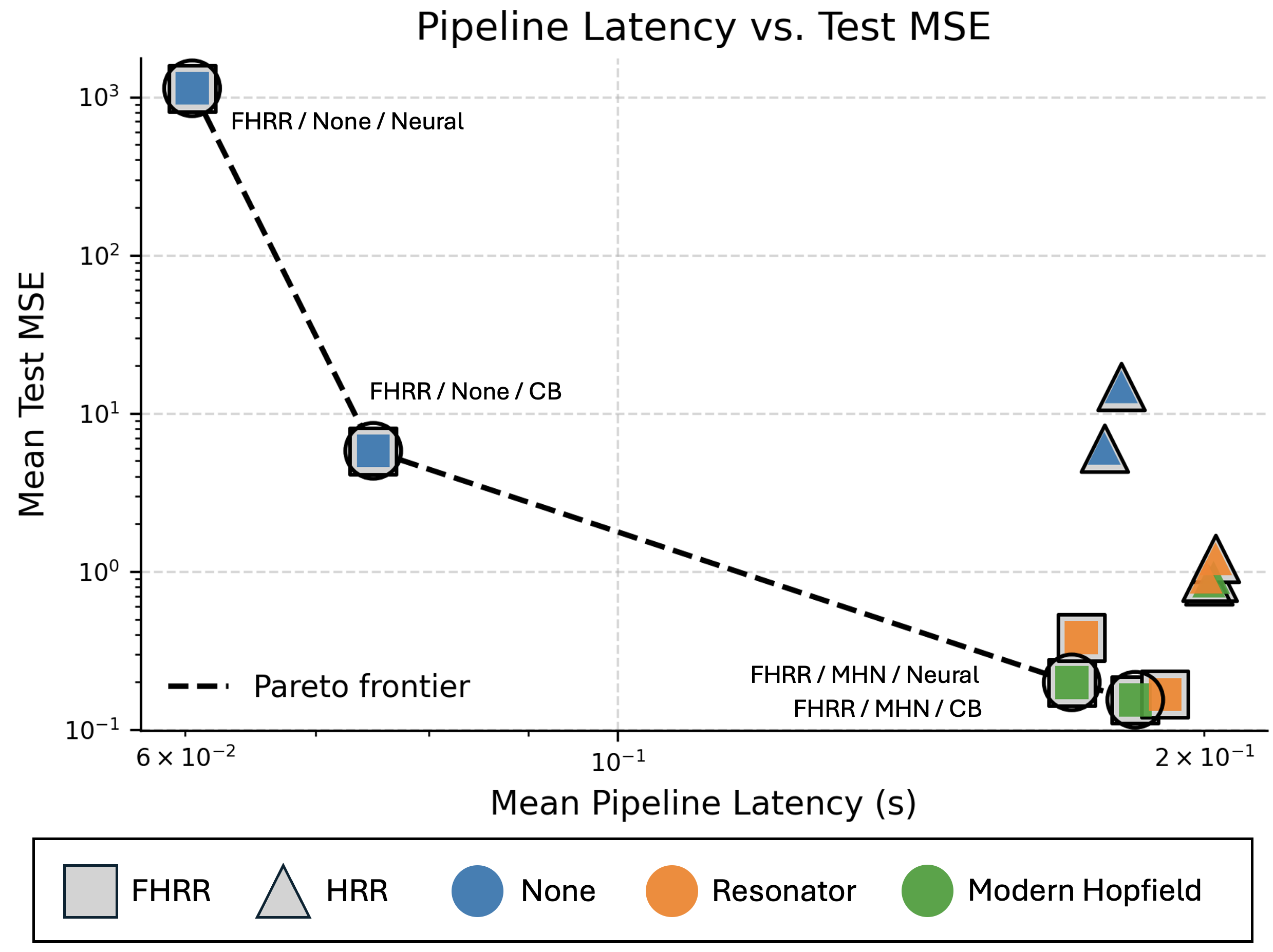}
    \caption{
        Latency–accuracy tradeoff across backends, cleanup, and regression methods. The x-axis shows the mean pipeline latency, while the y-axis shows the mean squared error. The dashed curve indicates the Pareto frontier.
    }
    \Description{
        Latency–accuracy tradeoff across backends, cleanup, and regression methods. The x-axis shows the mean pipeline latency, while the y-axis shows the mean squared error. The dashed curve indicates the Pareto frontier.
    }
    \label{fig:pareto}
\end{figure}

\begin{figure}
    \centering
    \includegraphics[width=0.7\linewidth]{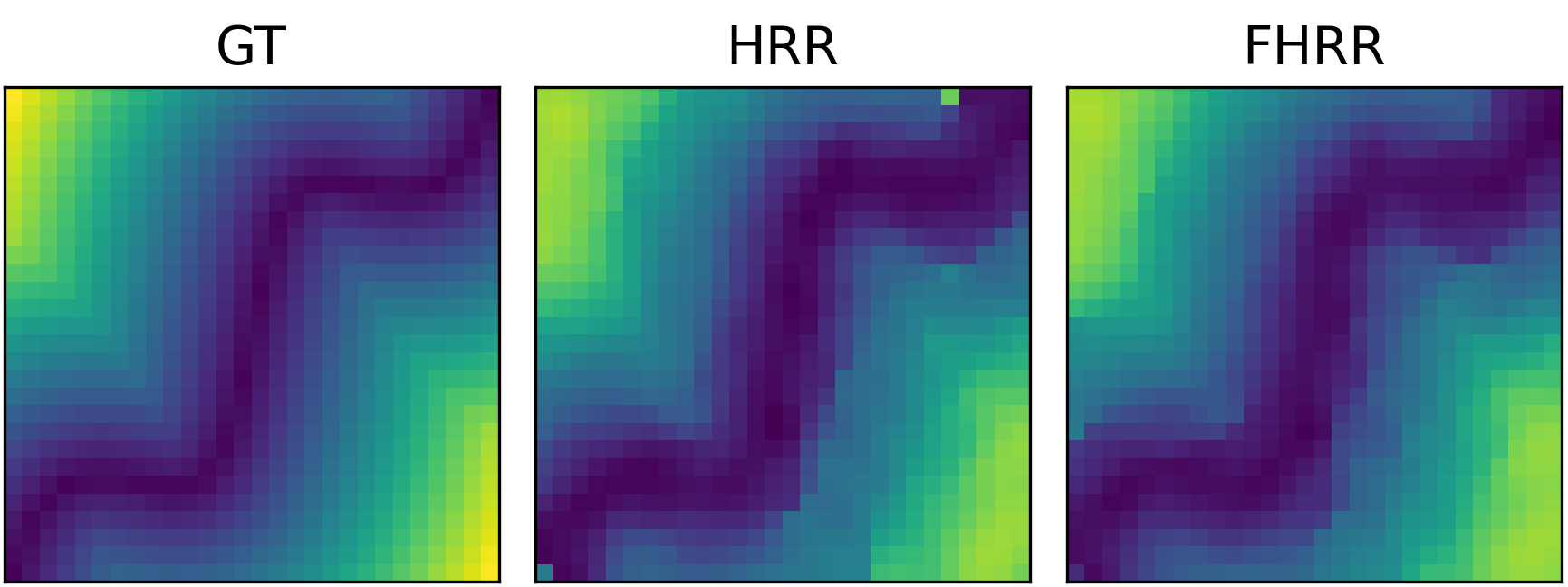}
    \caption{
        Comparison of reconstructions using HRR and FHRR. The left panel shows the ground truth generated from the environment. The middle and right panels show reconstructions produced by the best-performing HRR and FHRR configurations identified in the latency–accuracy analysis.
    }
    \Description{
        Comparison of reconstructions using HRR and FHRR. The left panel shows the ground truth generated from the synthetic environment. The middle and right panels show reconstructions produced by the best-performing HRR and FHRR configurations identified in the latency–accuracy analysis.
    }
    \label{fig:reconstructions}
\end{figure}

\vspace{5pt}
\noindent
\textbf{Experimental Setup:}
We evaluate HyperSpace with the following module and backend combinations: (i) HRR and FHRR backends, (ii) none, resonator, and modern Hopfield cleanup operations, (iii) codebook and neural network regression.
Latency is measured as the end-to-end runtime, in seconds, of the full pipeline on an Apple M4 Pro CPU.
To ensure a consistent comparison, all backends are evaluated using unoptimized implementations without hardware-specific acceleration.
As a result, the reported runtimes reflect baseline performance and may be improved with optimized kernels or specialized hardware.
Mapping performance is measured as the mean squared error (MSE). Each combination is evaluated across five random seeds with a vector dimensionality of 8096. We report the mean latency and prediction error for each combination.

\textbf{Pipeline Latency Breakdown:}
Figure~\ref{fig:pipeline-analysis} reports the mean latency of each combination with a breakdown across individual modules.
The stacked bars illustrate the relative contribution of positional encoding, value encoding, memory storage, positional inversion, cleanup, and regression.
Because HyperSpace decomposes this pipeline into abstract modules, these results reveal how different pipelines allocate computational effort and runtime latency.

Although FHRR provides the lowest overall latency, the latency distribution differs substantially between backends.
In HRR, encoding and memory storage contribute a larger fraction of the total runtime due to the FFT-based implementation of binding.
In contrast, FHRR performs these operations directly in the complex domain using element-wise multiplication.

Cleanup operations account for a larger share of runtime in FHRR configurations, likely due to the higher cost of complex-valued arithmetic relative to the real-valued operations used by HRR.
As a result, while HRR and FHRR exhibit comparable overall latencies, the allocation of computational effort across modules differs significantly.
This highlights a key strength of HyperSpace: its abstract modules expose discrepancies between isolated operators and end-to-end pipeline behavior.
It therefore reveals how backends reshape the computational profile of VSA-based spatial learning and introduce distinct trade-offs.

\textbf{Latency–Accuracy Trade-offs:}
Figure~\ref{fig:pareto} illustrates the trade-off between pipeline latency and reconstruction error across all combinations.
Each point corresponds to a unique combination of backend, cleanup, and regression strategies.
The dashed curve denotes the Pareto frontier, configurations that achieve optimal trade-offs between pipeline latency and reconstruction error.

Configurations without cleanup achieve the lowest latency but incur higher reconstruction error due to the noise introduced in the memory storage and positional inversion modules.
Conversely, resonator and modern Hopfield cleanup methods reduce reconstruction error by iteratively refining decoded vectors.
This process improves reconstruction fidelity but introduces additional computational overhead proportional to the codebook size and number of cleanup iterations.

The regression modules introduce additional trade-offs.
Codebook decoding relies on similarity comparisons against discretized entries in the codebook, enabling fast inference while constraining outputs to a finite set of predefined values.
Neural network regression instead learns a continuous mapping from vectors to scalar values, improving reconstruction fidelity while maintaining comparable or lower inference latencies.

However, this introduces additional considerations: neural regression requires offline training and reduces interpretability relative to explicit codebook-based decoding.
In our experiments, training takes approximately 10–20 seconds and is performed once.
As a result, the choice between regression methods reflects a trade-off between discrete, interpretable representations and continuous, learned mappings with potentially improved performance.

Together, these results highlight how the Pareto frontier emerges from competing factors across HyperSpace operators.
Rather than reflecting a single optimal backend, the frontier reveals a family of configurations that balance encoding efficiency, cleanup dynamics, reconstruction error, and memory complexity.

\textbf{Qualitative Reconstruction Analysis:}
Figure~\ref{fig:reconstructions} shows the reconstructions generated by the best-performing HRR and FHRR configurations in the latency–accuracy analysis.
The left panel shows the ground-truth map, while the middle and right panels show the reconstructions produced by HRR and FHRR, respectively.


Qualitatively, the HRR reconstruction exhibits slightly higher noise near boundaries and regions with steep gradients.
These artifacts are consistent with numerical effects introduced by FFT-based circular convolution~\cite{plate1991holographic}.
In contrast, FHRR performs these operations directly in the complex domain using element-wise complex multiplication.
This reduces the accumulation of numerical error during encoding and decoding~\cite{plate2003holographic}.

Despite these differences, both backends produce similar reconstructions and preserve the dominant spatial features.
This supports the earlier quantitative results showing that HRR and FHRR achieve comparable overall reconstruction performance, despite differences in how computational cost is distributed across the pipeline.
\section{Conclusion}
\label{section:conclusion} 

In this work, we introduced HyperSpace, a generalized framework for constructing, decomposing, and evaluating continuous spatial VSA pipelines through a shared set of abstract operators.
By formalizing spatial learning in terms of encoding, binding, bundling, inversion, similarity, cleanup, and regression, HyperSpace enables controlled end-to-end comparisons across heterogeneous backends and module configurations.
Therefore, HyperSpace serves not only as an implementation framework, but also as a tool for understanding how design choices propagate through the full pipeline.

Using this framework, we instantiated and analyzed representative HRR- and FHRR-based systems~\cite{plate1991holographic, plate2003holographic}.
Although FHRR provides lower theoretical complexity for several individual operations, these operator-level advantages did not directly translate into uniformly lower end-to-end cost.
Instead, HyperSpace revealed that similarity-driven cleanup and regression can dominate runtime in spatial learning, leading HRR and FHRR to exhibit comparable overall latency even though they rely on different operations.

HyperSpace also exposed deployment-level trade-offs that are not apparent from operator-level analysis alone.
HRR uses real-valued vectors and requires approximately half the storage of FHRR, while FHRR avoids FFT-based binding and inversion through direct complex-valued operations.
As a result, backend selection is not simply a question of asymptotic arithmetic complexity, but a broader system-level decision shaped by runtime distribution, memory footprint, implementation characteristics, hardware compatibility, and decoder design.

More broadly, these results illustrate the central value of HyperSpace: it provides a common framework for systematically analyzing spatial VSA systems as complete pipelines rather than isolated operators.
This makes it possible to study how backends, cleanup strategies, regression modules, and future extensions interact within end-to-end systems.
Future work will extend HyperSpace to support additional backends~\cite{Kleyko_2022}, provide hardware efficient backend implementations, hybrid and compressed representations~\cite{imani2019quanthd}, and larger applications in robotics~\cite{Renner2024, dumont2025symbols} and other resource-constrained environments~\cite{ni2025heal}.

\begin{acks}
We acknowledge the technical and financial support of the Automotive Research Center (ARC) in accordance with Cooperative Agreement W56HZV-19-2-0001 U.S. Army DEVCOM Ground Vehicle Systems Center (GVSC) Warren, MI.
\end{acks}

\bibliographystyle{ACM-Reference-Format}
\bibliography{sample-base}










\end{document}